\documentclass[letterpaper]{article}
\usepackage{aaai2026}  
\usepackage{times}                 
\usepackage{helvet}                
\usepackage{courier}               
\usepackage[hyphens]{url}          
\usepackage{graphicx}              
\usepackage{natbib}                
\usepackage{caption}               
\usepackage{booktabs}

\usepackage{algorithm}
\usepackage{algorithmic}
\usepackage{newfloat}
\usepackage{listings}
\DeclareCaptionStyle{ruled}{labelfont=normalfont,labelsep=colon,strut=off}
\floatstyle{ruled}
\newfloat{listing}{tb}{lst}
\floatname{listing}{Listing}

\title{Tri-Bench: Stress-Testing VLM Reliability on Spatial Reasoning\\
under Camera Tilt and Object Interference
}
\author{Amit Bendkhale}
\affiliations{
Independent Researcher\\
Pune, India\\
}

\begin{document}
\maketitle

\begin{abstract}
Verifiable geometric reasoning is a critical component for trustworthy and controllable agentic AI. Despite impressive capabilities, Vision-Language Models (VLMs) often fail under realistic scene changes. We present \emph{Tri-Bench}, a compact benchmark of planar triangle problems that isolates \emph{relative} geometric reasoning while stressing two deployment-critical factors: \emph{camera pose} (planar vs.\ tilted) and \emph{scene context} via object interference (10 everyday objects). To test verifiability and control, we evaluate four recent VLMs using a single, fixed prompt whose guardrail explicitly describes a surrounding square border, enabling correct answers via homography. We evaluate six simple tasks over binary and continuous targets, and observe that the overall accuracy with respect to 3D ground truth is modest, $\sim$69\% on average (best $\sim$75\%, worst $\sim$64\%). The same responses align even more closely with 2D projections in the image plane, where mean accuracy is $\sim$72\%. All four VLMs consistently fail, with accuracy falling to $\sim$0\%, on recognizing minority shape classes (equilateral, isosceles, right-angled triangles). Additionally, overall VLM accuracy degrades by $\sim$4.1\% under camera tilt. This demonstrates that models fail to correctly utilize the explicit frame-of-reference hint provided in the prompt and default to 2D image plane cues. Finally, we find that object interference has no significant effect on VLM accuracy.

\end{abstract}

\section{Introduction}
Recent advancements in agentic AI have made Vision-Language Models (VLMs) an integral part of real-world applications. Trustworthy and controllable agentic AI for robot navigation, AR/VR measurement tools, 3D reconstruction, AI-based 3D geometry teaching, and medical assistance depends heavily upon verifiable spatial reasoning. Though VLMs have shown impressive general visual reasoning capacities, their robustness in realistic geometric tasks remains a critical unverified barrier to deployment. Current benchmarks for spatial reasoning often focus either on estimating absolute distances, angles, orientation, etc.\ or on problem-solving in abstract diagrams or scenes. However, severe gaps exist to stress test against deployment-critical factors like camera pose invariance and object interference.

To address this fundamental gap, we present \emph{Tri-Bench}, a compact benchmark of 400 images built on camera captured planar triangle problems. We use the most fundamental closed geometric structures, namely triangles, for evaluating relative spatial reasoning instead of absolute values. Unlike absolute spatial reasoning, where identifying distances and angles matter, we focus on ratios of distances and differences of angles, which capture deeper spatial reasoning attributes in geometry. We provide a frame-of-reference guardrail to establish a pathway toward accurate geometric estimation using homography. Our results reveal significant robustness failures. Our primary contributions are: (i) A novel controlled benchmark, Tri-Bench, for diagnosing VLM spatial reasoning robustness to camera pose and object interference. (ii) A stress-test with comprehensive analysis of four leading VLMs, revealing a critical failure mode where models misinterpret the 3D real world as 2D projections in the image plane even when provided with sufficient guardrails. (iii) Identifying majority class bias in precision tasks (near-zero accuracy on minority shapes) with clear performance gaps across triangle shapes.

\section{Background and Related Work}
The evaluation of spatial reasoning in VLMs is a rapidly growing field. There exists significant work such as Mind the Gap \cite{mindthegap} and OmniSpatial \cite{omnispatial} on evaluation of broad, human-like cognitive skills. These test mental rotation, spatial visualization and navigation. Other works like iVISPAR \cite{ivispar} and "What's Up?" \cite{whatsup} probe the understanding of relative spatial relations (e.g., left/right, above/below). These benchmarks, even though foundational, often test broad cognitive logic rather than isolating robustness of fundamental geometric measurements. Tri-Bench complements this work by instead providing a narrow, deep, diagnostic probe for angles and distances under specific physical perturbations. 

Another major research area focuses on geometric reasoning from clean, symbolic inputs like text and diagrams, often in context of mathematical problem solving. Benchmarks such as MathBench \cite{mathbench} and VisioMath \cite{visiomath} evaluate VLMs on Olympiad-style geometry problems. NeSyGeo \cite{nesygeo} and AutoGPS \cite{autogps} are also powerful neuro-symbolic frameworks for data generation and deductive reasoning. Tri-Bench addresses the complementary challenge of reasoning from noisy, photorealistic images, bridging the gap between abstract logic and embodied real-world perception.

Our work is most closely related to benchmarks that test VLM robustness and specific spatial skills. DynaMath \cite{dynamath} tests robustness by introducing dynamic variants to math problems. SpatialVLM \cite{spatialvlm} focuses on improving spatial reasoning capabilities of VLMs and other works evaluate top-view reasoning \cite{topview} using spatial frames of reference \cite{representspace}. Tri-Bench builds directly on this as the first benchmark to systematically control and isolate the effects of camera pose and object interference on relative spatial reasoning in a controlled photorealistic setting. We therefore contribute a compact, reproducible benchmark where specific failure modes are interpretable.

\section{Dataset}

\subsection{Composition and Capture Conditions}
We use a 1 meter $\times$ 1 meter square and construct \textbf{100 labeled triangles} inside of it. This set of triangles is diverse with respect to shape and spatial orientation. There are \textbf{38 acute} triangles, \textbf{32 obtuse} triangles and, \textbf{30 right} triangles, with side types: \textbf{64 scalene}, \textbf{26 isosceles}, \textbf{10 equilateral}. No two triangles share the exact combination of shape, vertex labels and placement relative to the square.

\begin{figure}[t]
\centering
\includegraphics[width=\columnwidth]{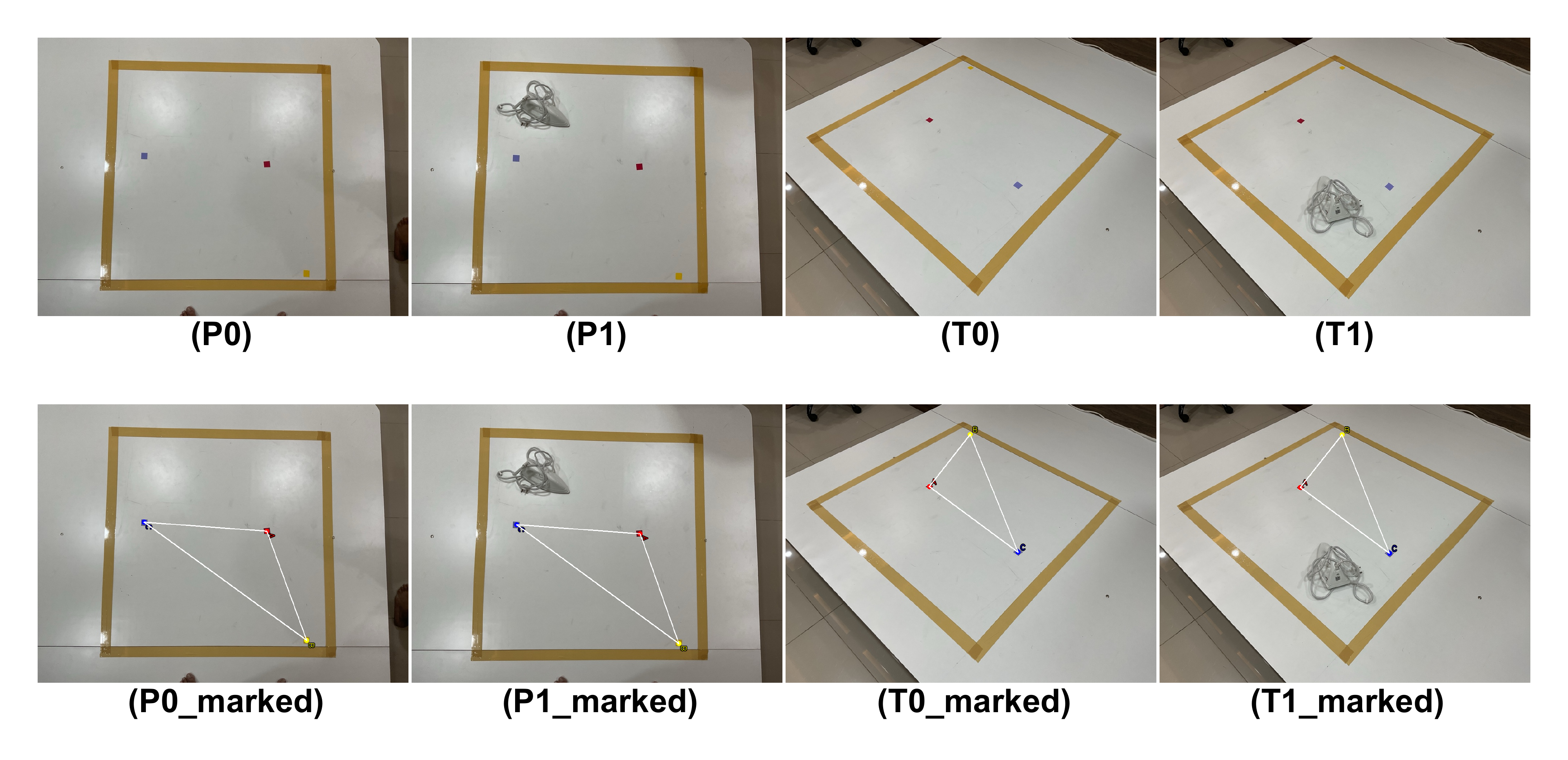}
\caption{Four capture conditions for triangle 037. Top row displays the four captures P0, P1, T0, and T1; bottom row displays the corresponding marked images. The triangle  is obtuse in P0\_marked but appears right-angled in T0\_marked.}
\label{fig:dataset_example}
\end{figure}

We captured \textbf{4 views} for each labeled triangle, as shown in Figure~\ref{fig:dataset_example}. We refer to these capture conditions as P0 (planar, no object), P1 (planar, with object), T0 (tilted, no object), and T1 (tilted, with object). For every captured image, we manually mark the pixel coordinates of points A, B, and C for measurements in the image plane. The dataset includes \textbf{95 uniquely shaped triangles} and \textbf{5 repetitions} to perturb labels and positions. Table~\ref{tab:confusion} quantifies how often these 3D shapes change type after 2D projection, for both side and angle classes. Across the 400 images, about 27\% of triangles change side type and 34\% change angle type between 3D and 2D labels. For side type, only $\sim$7\% of scalene images change class, versus 62.5\% for both isosceles and equilateral. For angle type, about 26\% of acute, 9\% of obtuse, and $\sim$70\% of right triangles change class in the image plane.

\subsection{Pose and Context Factors}
\textbf{Pose.} The \emph{planar} view approximates orthographic capture; the \emph{tilted} view introduces in-plane rotation and mild perspective (implicit scale change), with camera tilt angles varied by hand across a realistic range.

\textbf{Objects.}
We use ten everyday objects, each with 10 placements: Rubik's cube, glass vase, electric steam iron press, non-fiction book, water kettle, apple, make-up box, plastic stool, 15.6'' laptop, and a medium pillow. Objects are always placed inside the square border. Occlusions are annotated (25 images with partial occlusion of at least one triangle edge and exactly 1 image with a fully occluded triangle vertex); further capture details are given in Appendix~B.

\subsection{Labeling Rules and Tolerances}
Let (a, b, c) be side lengths sorted by magnitude. We define:
\begin{itemize}
    \item \textbf{Isosceles}: \\$\displaystyle \min\!\Big(\frac{|a-b|}{\max(a,b)}, \frac{|b-c|}{\max(b,c)}, \frac{|a-c|}{\max(a,c)}\Big) \le 3\%$.
    \item \textbf{Equilateral}: \\all pairwise relative side ratios lie within 3\%.
    \item \textbf{Right}: \\$\exists\,\theta \in \{\angle A,\angle B,\angle C\}$ s.t.\ $|\theta - 90^\circ| \le 2^\circ$.
\end{itemize}

All code, data, and scripts are available on GitHub.\footnote{\url{https://github.com/Amiton7/Tri-Bench}}. Additional construction details are provided in Appendix~B.

\begin{table}[t]
\centering
\small

\begin{minipage}{0.45\linewidth}
\centering
{\setlength{\tabcolsep}{2.5pt}%
\begin{tabular}{lccc}
\toprule
\textbf{2D$\backslash$3D} & Scal. & Isos. & Equi. \\
\midrule
Scal. & 237 & 65 & 20 \\
Isos. & 17  & 39 & 5  \\
Equi. & 2   & 0  & 15 \\
\bottomrule
\end{tabular}}
\caption*{(a) Side type}
\end{minipage}
\hspace{4pt}
\begin{minipage}{0.45\linewidth}
\centering
{\setlength{\tabcolsep}{2.5pt}%
\begin{tabular}{lccc}
\toprule
\textbf{2D$\backslash$3D} & Acute & Obtuse & Right \\
\midrule
Acute  & 113 & 8   & 35 \\
Obtuse & 34  & 116 & 49 \\
Right  & 5   & 4   & 36 \\
\bottomrule
\end{tabular}}
\caption*{(b) Angle type}
\end{minipage}

\caption{3D--2D shape label mismatch. Each cell gives the number of images whose triangle has a true 3D class (columns) and an apparent 2D class in the image plane (rows). For example, there are 65 images with real-world isosceles triangles that appear scalene in the  image plane.}

\label{tab:confusion}
\end{table}

\section{Tasks and Metrics}
We evaluate six well-rounded specific geometric reasoning tasks relative to the triangle formed by joining the points A, B and C in a given image, as below:
\begin{itemize}
    \item \textbf{Q1.} Is triangle ABC equilateral, isosceles, or scalene?
    \item \textbf{Q2.} Is triangle ABC acute, right, or obtuse?
    \item \textbf{Q3.} Estimate $AB/AC$.
    \item \textbf{Q4.} Estimate $|\angle ABC - \angle ACB|^\circ$.
    \item \textbf{Q5.} Estimate $\displaystyle \frac{\max\{AB,BC,CA\}}{\min\{AB,BC,CA\}}$.
    \item \textbf{Q6.} Estimate $\displaystyle
\max\!\bigl\{\,|\angle A-\angle B|,\,|\angle B-\angle C|,\,|\angle C-\angle A|\,\bigr\}\,^\circ$.
\end{itemize}

\noindent \textbf{Intuition:} Solving Q1 requires identifying if two planar line segments are equal without absolute measurements. Similarly, Q2 requires identifying the largest angle and comparing it with a right-angle. Solving Q3 and Q4 requires first point location, then comparing the "factor" by which one geometric entity (distance or angle) differs from the other. Q5 and Q6 require identifying the maximum and minimum geometric entities and computing the differentiating "factor". In all six questions, we do not provide exact distances, angles, or an absolute coordinate system. This forces the VLMs to use qualitative and quantitative reasoning for precise comparisons. Testing without such explicit measurements brings out the true capacity for geometric reasoning.

\subsection*{Unified Evaluation: Accuracy and Error}
For each task $t\in\{1,...,6\}$, the metrics are defined for a single image instance. We report accuracy $\kappa_{t}$ and error $\varepsilon_{t}$, where $\kappa_{t}=1-\varepsilon_{t}$. The  errors $\varepsilon_{t}$ are defined as below.

\noindent\textbf{Notation:} $\hat{y}=$ prediction made by VLM; $y=$ ground truth; $\theta_{max}=180^{\circ}$.

\noindent\textbf{Categorical (Q1, Q2).} The error is 1 if the prediction is wrong, 0 otherwise.
\[ \varepsilon_{t}=1-\mathbf{1}\{\hat{y}=y\}. \]

\noindent\textbf{Relative Ratio (Q3, Q5).} This is a relative error.
\[ \varepsilon_{t}=\min\Bigl(1, \frac{|\hat{y}-y|}{y}\Bigr). \]

\noindent\textbf{Normalized Angle (Q4, Q6).} The error is the absolute difference normalized by $\theta_{max}=180^{\circ}$.
\[ \varepsilon_{t}= \min \Bigl(1, \frac{|\hat{y}-y|}{\theta_{max}}\Bigr). \]

\noindent We use categorical error for tasks Q1 and Q2, to evaluate the precision of reasoning in shape classification. The relative errors for Q3 and Q5 are due to the fact that distance ratios are scale invariant (e.g. an error of 0.5  is more significant for true ratio of 1.5 than for true ratio of 5.0). Finally, for the angle tasks Q4 and Q6, normalized error bounds the metric to a range, allowing unified comparison across six tasks.

\section{Experimental Protocol}
\textbf{Models.} We evaluated four recent VLMs: Gemini 2.5 Pro, Gemini 2.5 Flash, GPT-5, and Qwen2.5-VL-32B. Prompts and parsing templates were identical for all models; sampling used API defaults. No chain-of-thought or structured rationales were used.

\textbf{Conditions.} We report accuracy ($\kappa$) separately for the four capture conditions P0, P1, T0, and T1. In the Results section, we also summarize brittleness by comparing planar vs.\ tilted (P0/P1 vs.\ T0/T1) and no object vs.\ with object (P0/T0 vs.\ P1/T1) accuracies.

\textbf{Prompting.} We evaluated all models using a single, fixed zero-shot prompt. The prompt defines the scene (e.g., A=RED), including the explicit "guardrail" hint of the "light-brown masking-tape square border." It then lists the six evaluation tasks and strictly enforces a JSON-only output with six pre-defined keys to ensure parsable responses. The full prompt is provided in Appendix~A.

\section{Results}

Our results reveal several critical failure modes in leading VLMs. First, we find that VLM estimates align more closely with the 2D projections in the image plane than the 3D ground truth, proving they fail to use the prompt's 'guardrail' hint. Second, accuracy on precision tasks (Q1/Q2) drops to almost 0\% for non-majority classes. Third, camera tilt consistently degrades performance, while object interference has a negligible effect. Across tasks, Gemini 2.5 Pro and Gemini 2.5 Flash consistently outperform GPT-5 and Qwen2.5-VL-32B.  Additionally, relative comparison tasks (Q4, Q6) are much easier for all models than absolute angle identification (Q2) and specific ratio estimation (Q3).

\subsection{3D vs. 2D Misinterpretation}
A core finding of our work is that VLMs are not performing true 3D-aware reasoning. As shown in Table~\ref{tab:3d_vs_2d}, when we re-compute accuracy with respect to the 2D projections in the image plane, the Gemini models gain roughly 5--6 \%, and the overall mean accuracy rises from 68.98\% (3D) to 72.32\% (2D). GPT-5 and Qwen2.5-VL-32B improve only slightly (by about 1--2\%). This strongly suggests that, on average, models default to 2D image-plane cues rather than using the 3D homography information provided by the surrounding square ``guardrail.''

\begin{table}[t]
\centering
\small
\begin{tabular}{lcc}
\hline
\textbf{Model} & \textbf{$\kappa_{3D}$} & \textbf{$\kappa_{2D}$} \\
\hline
Gemini 2.5 Pro   & 75.30 & 80.89 \\
Gemini 2.5 Flash & 71.58 & 77.14 \\
GPT-5        & 64.32 & 65.04 \\
Qwen2.5-VL-32B     & 64.70 & 66.22 \\
\hline
\textbf{AVERAGE} & \textbf{68.98} & \textbf{72.32} \\
\hline
\end{tabular}
\caption{Overall accuracy (\%) of VLMs w.r.t. 3D Ground Truth ($\kappa_{3D}$) and 2D projections in the Image Plane ($\kappa_{2D}$)}
\label{tab:3d_vs_2d}
\end{table}

\subsection{Precision Tasks and Majority Class Bias}
We find that VLMs are overconfident and fail dramatically at precision tasks. Table \ref{tab:precision_failure} shows the breakdown of the accuracy of Q1 and Q2 by the \emph{true} shape. While accuracy for 'Scalene' (99.51\%) and 'Acute' (85.69\%) appears high, this is deceptive. These are the majority classes in our dataset (64\% and 38\% respectively). For the rarer, precision-critical classes, accuracy plummets to \textbf{1.88\% (Right-angled)}, \textbf{1.44\% (Isosceles)}, and \textbf{0.00\% (Equilateral)}. This suggests VLMs are not performing fine-grained reasoning but are exhibiting a strong \textbf{majority class bias}, defaulting to "Scalene" or "Acute" regardless of the image's true geometry. Qwen2.5-VL-32B shows a total failure, scoring 0\% on all non-majority shape classes.

\begin{table}[t]
\centering
\small
\setlength{\tabcolsep}{3pt}
\begin{tabular}{lcccccc}
\hline
\textbf{Model} & \textbf{Scal.} & \textbf{Isos.} & \textbf{Equi.} & \textbf{Acute} & \textbf{Obtuse} & \textbf{Right} \\
& (Q1) & (Q1) & (Q1) & (Q2) & (Q2) & (Q2) \\
\hline
Gemini 2.5 Pro   & 99.61 & 2.88 & 0.00 & 78.29 & 88.28 & 0.00 \\
Gemini 2.5 Flash & 98.83 & 1.92 & 0.00 & 72.37 & 80.47 & 5.83 \\
GPT-5        & 99.61 & 0.96 & 0.00 & 92.11 & 3.91 & 1.67 \\
Qwen2.5-VL-32B     & 100.00 & 0.00 & 0.00 & 100.00 & 0.00 & 0.00 \\
\hline
\textbf{AVERAGE} & \textbf{99.51} & \textbf{1.44} & \textbf{0.00} & \textbf{85.69} & \textbf{43.16} & \textbf{1.88} \\
\hline
\end{tabular}
\caption{Accuracy (\%) on precision tasks (Q1, Q2)  drops significantly to near-zero for all non-majority shape classes, revealing a strong majority class bias.}
\label{tab:precision_failure}
\end{table}

\subsection{Effect of Tilt and Object Interference}
\begin{figure}[t]
\centering
\includegraphics[width=\columnwidth]{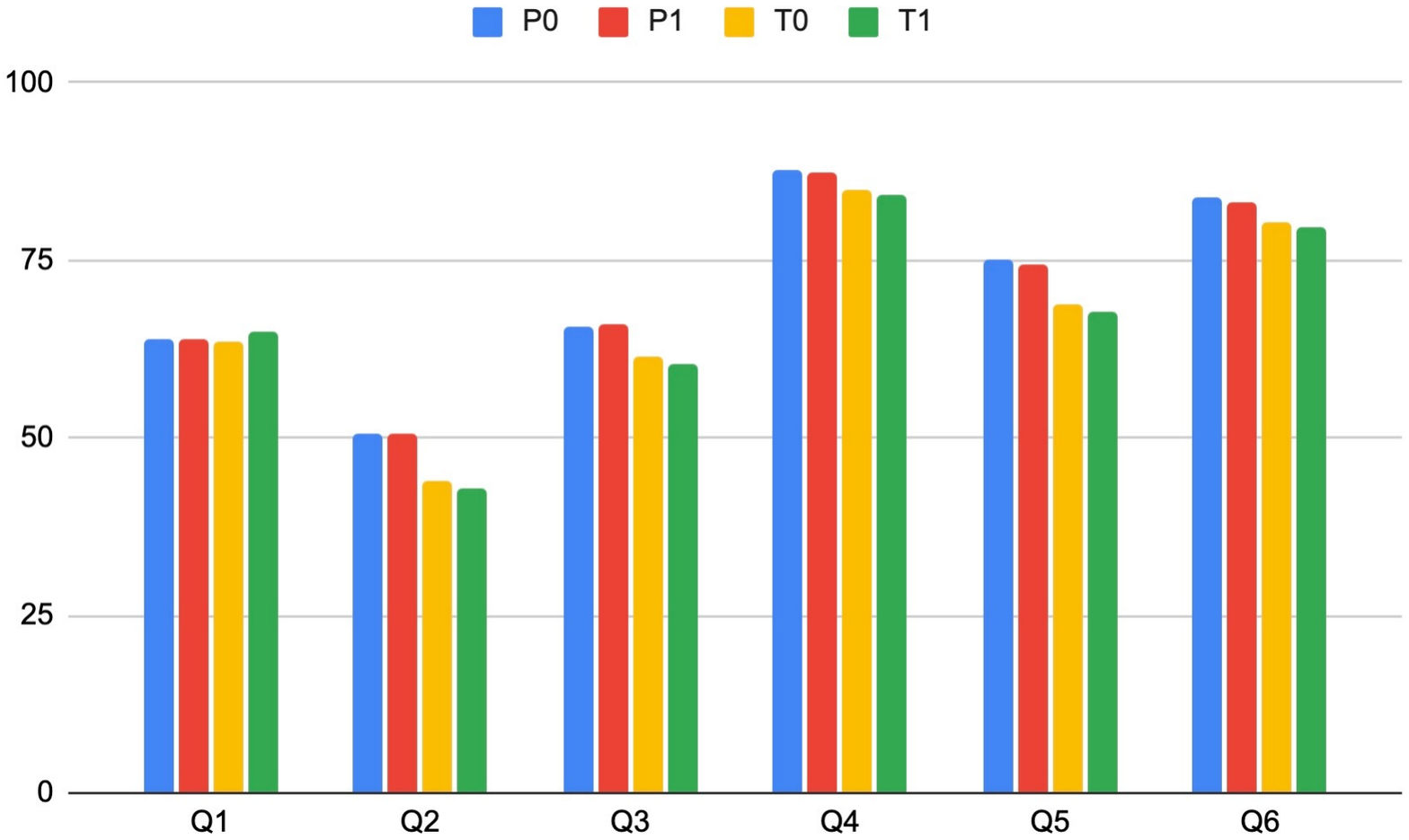}
\caption{Average accuracy across all models for each task (Q1--Q6) under the four capture conditions P0, P1, T0, and T1. Tilted views (T0/T1) are consistently less accurate than planar views (P0/P1), while the presence of an object (P1/T1 vs.\ P0/T0) has only a minor effect.}

\label{fig:main_graph}
\end{figure}

Figure~\ref{fig:main_graph} shows average accuracy across all four models for each task and capture condition.

\textbf{1. Camera tilt degrades performance.}
Across all models and questions, planar views (P0/P1) reach \textbf{71.0\%} accuracy, while tilted views (T0/T1) drop to \textbf{66.9\%}, i.e. about \textbf{4\%}. For Q2, and Q5, tilt reduces accuracy by roughly \textbf{6--7 \%}, and for Q3 by about \textbf{5\%} indicating a clear lack of pose invariance.

\textbf{2. Object interference is negligible.}
Aggregating over pose and question, ``no-object'' images (P0/T0) achieve \textbf{69.2\%} accuracy versus \textbf{68.8\%} for ``with-object'' images (P1/T1), i.e., a difference of less than \textbf{1\%}. This suggests that VLMs are relatively robust to this form of contextual clutter.

\subsection{Overall Model and Task Performance}
Across the benchmark, \textbf{Gemini 2.5 Pro} is the strongest model, with \textbf{Gemini 2.5 Flash} close behind; \textbf{GPT-5} and \textbf{Qwen2.5-VL-32B} consistently achieve lower accuracy (Table~\ref{tab:3d_vs_2d}). For Q1, overall accuracy is 64.06\%, almost identical to the proportion of scalene triangles in the dataset (64\%), and remains near 64\% across all four capture conditions (P0-T1), indicating a strong, pose- and context-independent majority class bias (Table~\ref{tab:precision_failure}). Figure~\ref{fig:main_graph} reveals a clear performance gap between task types: within angular reasoning, \textbf{absolute angle-type identification (Q2)} is much harder than \textbf{relative angle comparison (Q4, Q6)}, and the max/min side ratio task \textbf{Q5} is also noticeably easier than the absolute ratio estimation task \textbf{Q3}.

\section{Limitations and Future Work}

Our study is subject to several key limitations. The benchmark's construction is controlled, with all triangles always co-planar with the square border and fixed lighting, which might not capture the full variance of real-world captures. 

Our evaluation is also based on a single image. Inference through multi-view geometry seems a natural extension, with minimal guardrails. 

While we test 'tilt' as a binary factor (planar vs. tilted), a more granular study could measure the precise relationship between specific tilt angles and accuracy degradation. 

We deliberately use a single fixed guardrail prompt to test whether VLMs can follow an explicit frame-of-reference. More advanced prompting can be done in the future.

Though our metrics address a significant gap in existing literature, there still exist other gaps, where different evaluation metrics can be followed on our dataset.

It is possible that the majority class bias shown in Table \ref{tab:precision_failure} is  inherited from real-world training data, where scalene and acute triangles are more common, but we do not analyze training distributions and leave this to future work.

Finally, our work is limited to triangles. A necessary extension will be to apply this methodology to more complex shapes, such as other polygons, self-intersecting curves, and smooth differentiable surfaces.

\section{Conclusion and Broader Impact}

We introduced Tri-Bench, a compact benchmark for VLM spatial reasoning that isolates the effects of camera pose and object interference. We identified critical failure modes, finding that VLMs default to 2D image plane reasoning and fail to use explicit 3D "guardrail" prompts. This 2D bias leads to a strong majority class bias, with precision accuracy on non-majority shapes dropping to near-zero, and a consistent performance degradation under camera tilt.

These findings are significant for the "Trustworthy Agentic AI" community. They demonstrate that verifiability and control are not guaranteed by simple prompting. If an agent cannot be trusted to perform basic 3D reasoning on a simple triangle, it cannot be deployed in safety-critical robotics or autonomous navigation tasks where such reasoning is paramount. Tri-Bench offers a minimal, reproducible diagnostic for this critical capability gap, marking a necessary step toward building genuinely robust, controllable, and trustworthy agentic AI.

\bibliography{refs}

\appendix
\section{Prompt and Model Details}

\subsection{Exact Text Prompt}

\begin{quote}\small
The image shows triangle ABC whose vertices are the centres of three small coloured
square stickers: A=RED, B=YELLOW, C=BLUE.
A light-brown masking-tape square border surrounds the scene in the same plane as
triangle ABC. All questions refer to triangle ABC. Angles are in DEGREES.
$\angle ABC$ denotes the interior angle at vertex B.
Round all numeric answers to EXACTLY 4 decimals.

Q1. Is triangle ABC equilateral, isosceles, or scalene?\\
Q2. Is triangle ABC acute, right, or obtuse?\\
Q3. In triangle ABC, by what factor is the length AB greater than length AC?
(i.e., estimate AB / AC.)\\
Q4. In triangle ABC, by how much do angles $\angle ABC$ and $\angle ACB$ differ?
(i.e., estimate $|\angle ABC - \angle ACB|$ in degrees.)\\
Q5. In triangle ABC, what is (longest side) / (shortest side)?\\
Q6. In triangle ABC, what is (largest interior angle -- smallest interior angle)
in degrees?\\
Return STRICT JSON ONLY -- no prose, markdown, code fences, or extra keys. Use
EXACTLY these keys; numbers must have exactly 4 decimals:\\[2pt]
\texttt{"side\_type"}: one of ``equilateral'', ``isosceles'', ``scalene''.\\
\texttt{"angle\_type"}: one of ``acute'', ``right'', ``obtuse''.\\
\texttt{"ab\_over\_ac"}: number with 4 decimals.\\
\texttt{"abs\_b\_minus\_c\_deg"}: number with 4 decimals.\\
\texttt{"max\_over\_min\_side"}: number with 4 decimals.\\
\texttt{"angle\_range\_deg"}: number with 4 decimals.\\
Output only the JSON object with these six keys and the computed values for THIS image.
\end{quote}

\subsection{Evaluated VLMs and API Identifiers}

Table~\ref{tab:vlm-apis} lists the exact API names used in our experiments. The open-source model Qwen2.5-VL-32B was accessed through an API on fireworks.ai. For GPT-5 (vision), we query via OpenAI's \texttt{gpt-5-chat-latest} endpoint, which at the time of our experiments resolved to snapshot \texttt{gpt-5-2025-08-07}.

\begin{table}[t]
\centering
\small
\begin{tabular}{ll}
\toprule
\textbf{VLM} & \textbf{API identifier} \\
\midrule
Gemini 2.5 Pro      & \texttt{gemini-2.5-pro-preview-03-25} \\
Gemini 2.5 Flash    & \texttt{gemini-2.5-flash} \\
GPT-5      & \texttt{gpt-5-2025-08-07} \\
Qwen2.5-VL-32B      & \texttt{qwen2p5-vl-32b-instruct} \\
\bottomrule
\end{tabular}
\caption{VLMs and API identifiers used in all experiments.}
\label{tab:vlm-apis}
\end{table}


\section{Dataset Construction}

We built the dataset manually in a realistic indoor setting.  All side lengths were measured with a regular measuring tape. We ensured that the interior square border is exactly 100 cm $\times$ 100 cm (width of tape = 4.8 cm). Three square-shaped (3 cm $\times$ 3 cm) thin coloured paper cut-outs were used for labeling points A, B, and C.   Before taking any photos, we first  fixed the target triangle shapes to cover a diverse set of side--angle patterns, and then varied the position of each triangle on the flat surface so that shape and location are not tied together. For equilateral and right-angled triangles we used a protractor with fine markings. For isosceles triangles we first placed two vertices and then positioned the third so that  one of the new sides matched in length. Obtuse and acute triangles were constructed using our own simple geometric insights.

We aimed for a $4\!:\!3\!:\!3$ split of acute/obtuse/right and a $6\!:\!3\!:\!1$ split of scalene/isosceles/equilateral. Because final labels are assigned from precise measurements, a few manually placed triangles that were intended to be acute ended up just inside the obtuse or right categories, and a few intended isosceles triangles moved into the scalene class by a very small margin. We also pre-planned the pairing of objects and triangles so that each object interferes with exactly 10 triangles (20 images across P1 and T1) and appears across different shape and angle types (for example, the ``apple'' object is used with acute, obtuse, and right triangles).

Lighting came from 9 ceiling lights with curtains closed to avoid daylight. We tried our best to keep any object shadows outside the taped square border. Images were captured using the default camera app on an iPad Air (M1, 2022); the original \texttt{.HEIC} files were converted to \texttt{.JPG} at a fixed resolution using a simple Python script. Camera tilt was varied by hand to mimic realistic usage, matching practical scenarios where VLM-powered agents might be used for everyday measurement tasks. Partial occlusion of the square border or a triangle edge was sometimes unavoidable for larger objects, especially when we varied camera tilt. Even counting very small or tangential overlaps, this affects only about 6.25\% of all images. We intentionally kept exactly 1 image in which a triangle vertex is completely occluded. These occlusion cases add an extra robustness factor to the dataset.

\begin{figure}[t]
\centering
\includegraphics[width=0.67\columnwidth]{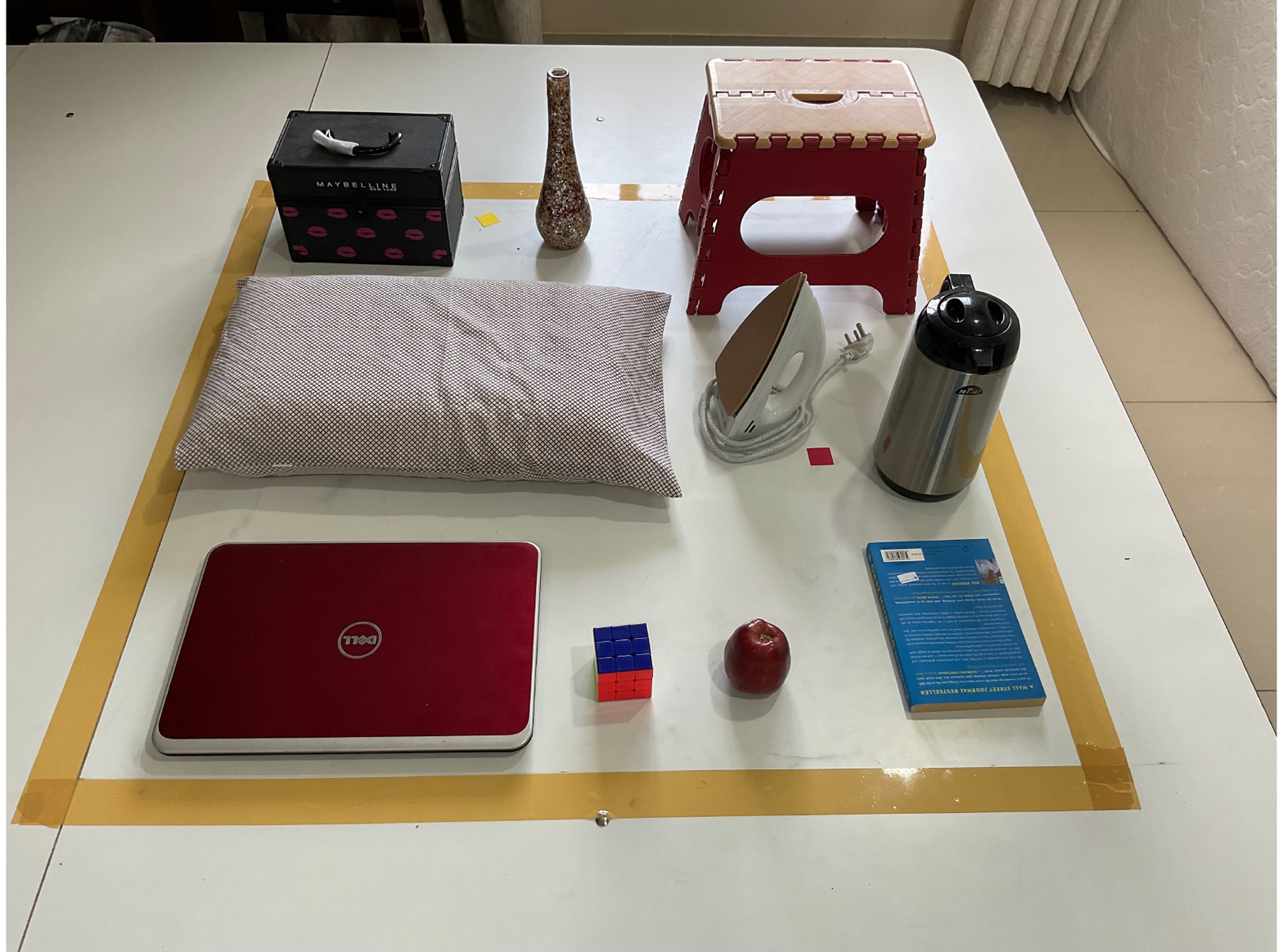}
\caption{The ten everyday objects used in Tri-Bench.}
\label{fig:objects}
\end{figure}

The choice of objects was meant to probe different kinds of interference. We selected ten everyday items varying in rectilinearity (flat-sided objects versus more organic shapes), familiarity (how easy it is to guess their size), and scale (small items such as a Rubik's cube versus larger items such as a pillow). Figure~\ref{fig:objects} shows all ten objects in a single image (captured under slightly different lighting).

\section{Homography-Based Estimation}

Because every triangle in Tri-Bench lies in the same plane as the taped
square border, a single homography per image is enough to recover the
triangle geometry (up to a global similarity). Conceptually, one simple procedure is:

\begin{enumerate}
    \item Extract the image-plane pixel coordinates of the four tape corners and sticker centres of A, B, and C.
    \item Estimate the $3\times3$ homography matrix $H$ that maps the tape's corners to a canonical unit square with vertices $(0,0)$, $(0,1)$, $(1,1)$, and  $(1,0)$ (in clockwise order).
    \item Apply $H$ to the image-plane coordinates of A, B, and C to obtain their coordinates in this normalized space.
    \item In the normalized space, compute side lengths and interior angles of triangle ABC, and the answers to Q1-Q6.
\end{enumerate}

Since $H$ is determined up to a global similarity transform (rotation,
translation, and uniform scale), all angle measurements and all ratios
of side lengths are preserved. This is sufficient for our evaluation tasks, which
depend only on shape types, length ratios, and angle differences
rather than on absolute metric units. A short script implementing this
procedure for triangle 037 (T0) is included in our GitHub repository.\footnote{\url{https://github.com/Amiton7/Tri-Bench}}

\end{document}